\documentclass[letterpaper,10pt,conference]{IEEEtran}
\IEEEoverridecommandlockouts
\usepackage[T1]{fontenc}
\usepackage{amsmath,amssymb}
\usepackage{booktabs}
\usepackage{graphicx}
\usepackage{multirow}
\usepackage[hidelinks]{hyperref}
\usepackage{cite}
\usepackage{placeins}

\usepackage[T1]{fontenc}
\usepackage{mathptmx}

\newcommand{\method}{TM-APR}
\newcommand{\tap}{TM-APR} 
\newcommand{\figplaceholder}[1]{\fbox{\rule{0pt}{#1}\rule{0.94\linewidth}{0pt}}}

\title{\bf\Large%
TM-APR: Thermal Temporal-Memory Localization via Analytic Online Adaptation
}
\author{Yanshuo Bai ~~~~~~~~~ Kanji Tanaka}

\begin{document}
\maketitle

\begin{abstract}
Thermal Visual Place Recognition (Thermal VPR) maps camera observations to metric poses within a mapped environment, serving as a prerequisite for autonomous navigation. 
However, thermal VPR suffers from severe environmental dependence, heavy online retraining overheads, and an inability to model dynamic non-linear shifts, causing existing frameworks to fail during online deployment. 
To achieve robust domain-invariant place recognition, we bridge Analytic Class-Incremental Learning (ACIL) with domain-invariant VPR for the first time, revealing that its gradient-free matrix updates construct a surprisingly strong baseline that outperforms conventional fine-tuning. 
Nevertheless, standard ACIL exhibits a critical vulnerability to extreme non-linear thermal fluctuations due to its structural linear assumptions. 
To overcome this limitation, we exploit a novel algebraic equivalence between ACIL and modern control theory, proposing \method{} which embeds Unscented propagation (U-ACIL), Gaussian Mixture partitioning (GMM-ACIL), and minimax $H_\infty$ optimization ($H_\infty$-ACIL) directly into the update loop. 
Our formulation guarantees exact closed-form matrix updates within $\mathcal{O}(1)$ computational complexity, bypassing backpropagation to ensure that the online update latency ($\Delta t_{\mathrm{learn}}$) remains strictly bounded below the sensor acquisition interval ($\Delta t_{\mathrm{acquire}}$), thereby eliminating trajectory jumps in real-time SLAM pipelines.
\end{abstract}

\section{Introduction}
Thermal Visual Place Recognition (Thermal VPR) maps camera observations to metric poses within a mapped environment, serving as a prerequisite for autonomous navigation \cite{thermalloc, anythermal}.

However, thermal VPR suffers from severe environmental dependence, heavy online retraining overheads, and an inability to model dynamic non-linear shifts. Thermal surface radiance distributions fluctuate continuously across temporal, seasonal, and weather transitions. These domain shifts distort latent representations, inducing false-positive retrievals that cause trajectory jumps. Existing thermal VPR frameworks predominantly rely on heavy offline batch training or fine-tuning foundation backbones \cite{thermalloc, anythermal}, which fail to adapt to dynamic non-linear thermal shifts during online deployment.

To achieve robust domain-invariant place recognition, we establish an Analytic Class-Incremental Learning (ACIL) \cite{acil} framework tailored specifically for thermal VPR. Departing from iterative gradient descent---which suffers from online overfitting, feature collapse, and latency---ACIL analytically maintains optimal feature statistics via closed-form matrix updates. By accumulating autocorrelation and cross-correlation matrices, this mechanism reveals a surprisingly strong domain-invariant baseline that outperforms conventional gradient-based fine-tuning and state-of-the-art baselines. To our knowledge, this work is the first to bridge ACIL with domain-invariant VPR.

Nevertheless, standard ACIL exhibits a critical vulnerability to extreme non-linear thermal fluctuations. Relying on linear analytical updates, severe thermal domain shifts violate its structural assumptions, degrading the latent geometry and reintroducing retrieval ambiguity in dynamic environments.

\begin{figure}[t]
\centering
\hspace*{-5mm}\includegraphics[width=9cm]{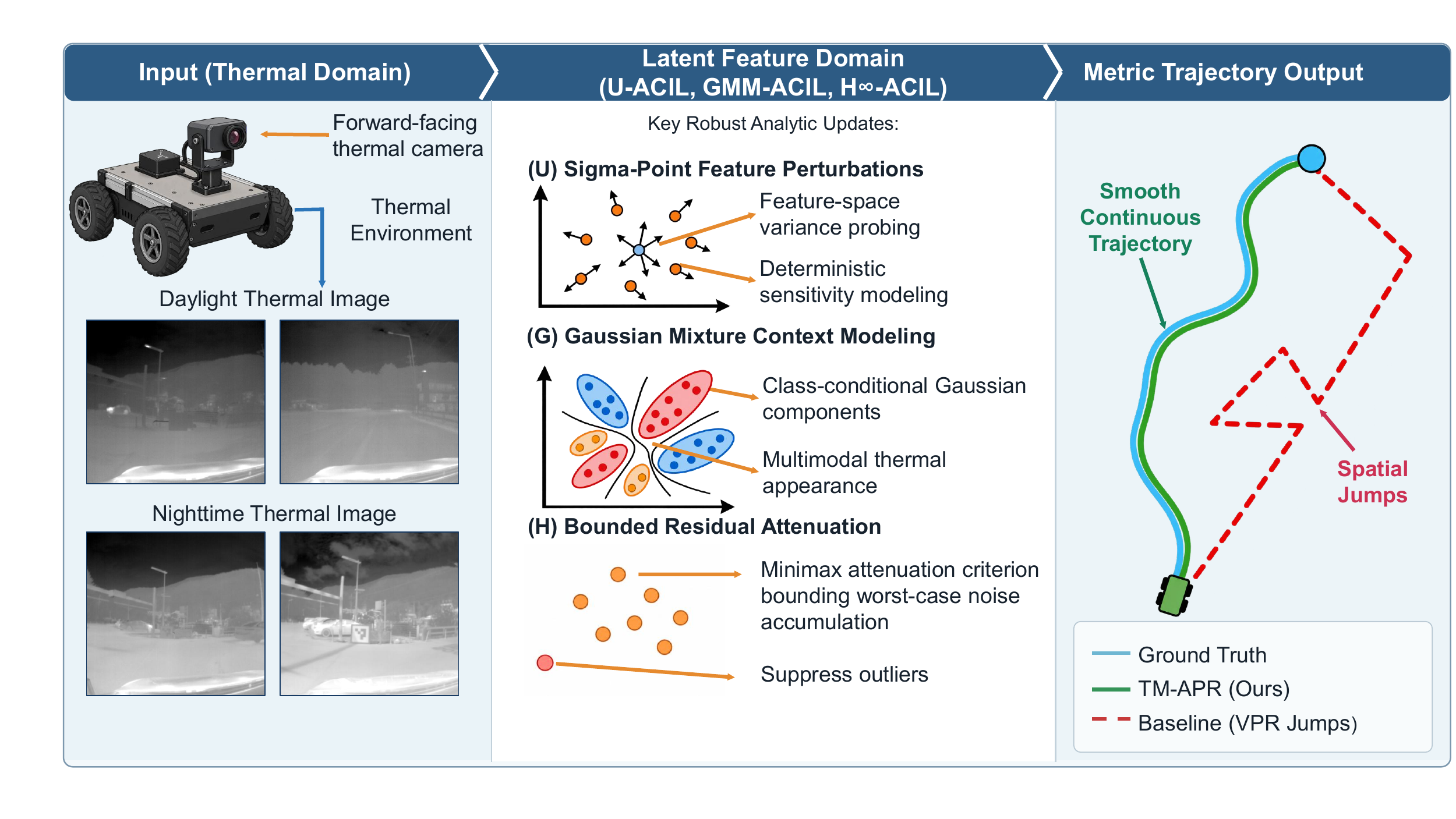}\vspace*{-5mm}\\
\caption{\textbf{TM-APR couples online thermal place recognition with recursive analytic updates ($U$-, GMM-, and $H_\infty$-ACIL) in the latent space.} By resolving non-linear domain shifts and rejecting ambiguous thermal retrievals without backpropagation, TM-APR eliminates spatial trajectory jumps (red dashed line) from baseline VPR, yielding a continuous and precise metric pose estimate (green solid line).}
\label{fig:teaser}
\end{figure}

To overcome this limitation, we exploit a novel algebraic equivalence between ACIL and modern control theory to capture complex non-linearities. Specifically, by mapping ACIL's Woodbury-based matrix updates to state estimation, we propose \method{} (Fig.~\ref{fig:teaser}), which embeds robust non-linear extensions directly into ACIL's analytical update loop. We introduce Unscented perturbation propagation (U-ACIL) to isolate local non-linearities, Gaussian Mixture space partitioning (GMM-ACIL) to suppress multi-modal variations, and minimax $H_\infty$ optimization ($H_\infty$-ACIL) to bound worst-case noise propagation for mathematically grounded online adaptation.

To satisfy real-time robotics constraints, our formulation inherently guarantees exact closed-form matrix updates within $\mathcal{O}(1)$ computational complexity. Bypassing backpropagation, the online adaptation latency ($\Delta t_{\mathrm{learn}}$) remains strictly bounded well below the sensor acquisition interval ($\Delta t_{\mathrm{acquire}}$). This fixed-time update property enables seamless onboard execution and eliminates trajectory jumps in real-time SLAM systems under continuous environmental transitions.

Our primary contributions are summarized as follows:
\begin{enumerate}
\item We discover that recursive ACIL provides a surprisingly strong domain-invariant baseline for thermal VPR, opening a new direction for online thermal localization.
\item We reveal a structural algebraic equivalence between ACIL's Woodbury-based updates and state estimation, and propose \method{}---unifying U-ACIL, GMM-ACIL, and $H_\infty$-ACIL---to resolve thermal feature non-linearities and suppress false positives.
\item We exploit the inherent $\mathcal{O}(1)$ closed-form matrix update of \method{} as a key byproduct, satisfying the strict real-time incremental learning demands of SLAM systems without backpropagation.
\item We validate the complete localization pipeline across three thermal benchmarks (MS2, STherO Valley, and a physical robot dataset), demonstrating state-of-the-art retrieval accuracy, trajectory continuity, and real-time efficiency.
\end{enumerate}

\section{Related Works}
We review thermal visual localization and analytic learning to motivate \method{} by clarifying the structural limitations of standard analytic methods under dynamic thermal conditions.

\subsection{Thermal Visual Place Recognition and Localization}\label{sec:2a}
Thermal VPR and Absolute Pose Regression (APR) suffer from appearance variations, low contrast, and thermal ambiguities. Early visual localization explored uncertainty modeling~\cite{kendall2016modelling}, temporal LSTMs~\cite{walch2017image}, geometric constraints~\cite{brahmbhatt2018geometry}, attention~\cite{atloc}, DarkLoc+~\cite{zhou2024darkloc}, and transformers~\cite{shavit2021learning}. Early thermal works addressed non-uniformity correction~\cite{borges2016practical}, thermal-inertial alignment~\cite{khattak2019keyframe}, and stereo tracking~\cite{mouats2015thermal}.

For spatial features, attention~\cite{atloc} and low-light constraints~\cite{zhou2024darkloc} complemented EfficientNet~\cite{tan2019efficientnet} and ViT~\cite{dosovitskiy2020vit} backbones. ThermalLoc~\cite{thermalloc} combined EfficientNet and ViT for thermal features, while subsequent work targeted driving scenarios~\cite{robustloc}. Recently, AnyThermal~\cite{anythermal} distilled visual foundation models (e.g., DINOv2). However, these static models cannot adapt online to dynamic thermal variations due to frozen place representations. Consequently, global descriptors remain vulnerable to non-linear noise and spatial overlaps, inducing overconfident false positives and trajectory jumps.

\subsection{Analytic Incremental Learning in Machine Learning}\label{sec:2b}
Analytic Class-Incremental Learning (ACIL)~\cite{acil} offers a gradient-free paradigm to prevent catastrophic forgetting and latency without raw data storage. Grounded in recursive least-squares (RLS)~\cite{ref72}, ACIL updates classifier boundaries in closed form, matching joint batch-training performance. ACIL has been extended across domains: Gaussian kernels (GKEAL~\cite{zhuang2023gkeal}), dual-stream compensation (DS-AL~\cite{zhuang2024dsal}), unexposed categories (GACL~\cite{zhuang2024gacl}), time series (TS-ACL~\cite{zhuang2024tsacl}), class-imbalanced driving (AEF-OCL~\cite{zhuang2025aefocl}, AIR~\cite{fang2024air}), and federated learning (DeepAFL~\cite{tang2026deepafl}). These methods achieve exact joint-learning equivalence with zero backpropagation and strict $\mathcal{O}(1)$ complexity.

\subsection{Motivation and Key Insight}
Although recursive ACIL provides a potent baseline for thermal VPR, standard ACIL relies on linear ridge regression over frozen features. Under continuous temperature shifts, this linear formulation struggles to fully capture non-linear feature overlaps, multi-modal shifts, and thermal noise.

Our key insight bridges this gap: discovering an algebraic equivalence between ACIL's Woodbury-based update matrix and classical state estimation reveals that the performance degradation in standard ACIL under thermal shifts structurally mirrors state estimation under non-Gaussian noise. Driven by this connection, \method{} incorporates three control-theoretic modules directly into ACIL's update loop: U-ACIL (Unscented perturbation propagation for local non-linearities), GMM-ACIL (Gaussian Mixture partitioning for multi-modal shifts), and $H_\infty$-ACIL (minimax $H_\infty$ criteria to bound worst-case noise). This unification resolves dynamic thermal non-linearities while preserving strict $\mathcal{O}(1)$ streaming adaptation without backpropagation.

\begin{figure*}[t]
\centering
\includegraphics[width=\textwidth]{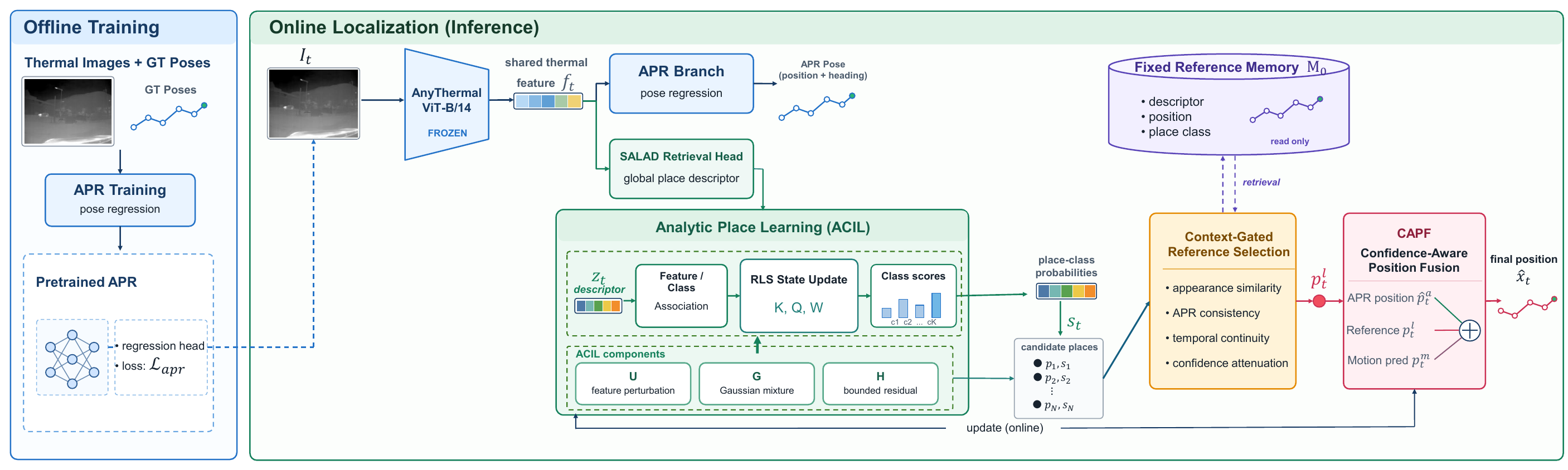}\vspace*{-5pt}\\
\caption{\textbf{Overview of the proposed \method{} pipeline.} Designed for real-time, lightweight thermal localization, \method{} couples continuous absolute pose regression (APR) with selective reference-place corrections without backpropagation. A frozen thermal representation feeds both a planar APR head and a place-retrieval branch. The online classifier state is decoupled from coordinate-bearing reference memory, enabling real-time adaptation via recursive analytic updates. Control-theoretically grounded extensions (U-, GMM-, and $H_\infty$-ACIL) resolve thermal non-linearities and overconfident false positives while strictly preserving $O(1)$ memory complexity. Confidence-Aware Position Fusion (CAPF) filters ambiguous retrievals to seamlessly integrate reference observations into a smooth, jump-free trajectory estimate.}
\label{fig:overview}
\end{figure*}

\section{Method}

\subsection{Problem Formulation and Information Flow}
As shown in Fig.~\ref{fig:overview}, \method{} processes an incoming thermal image \(I_t\) at time \(t\) to estimate a planar position \(\hat{\mathbf p}_t\in\mathbb R^2\). The APR branch is supervised by training poses. The reference traversal provides a fixed memory \(\mathcal M_0=\{(\mathbf d_j,\mathbf p_j,c_j)\}_{j=1}^{N}\), where \(\mathbf d_j\) is a normalized descriptor, \(\mathbf p_j\) its recorded position, and \(c_j\) its spatial place label (where multiple images can share a single class).

We explicitly decouple \(\mathcal M_0\) from the online adaptation state \(\mathcal A_t\), which includes classifier weights, sufficient statistics, and class-conditional models. During localization, \(\mathcal M_t=\mathcal M_0\) remains fixed, whereas \(\mathcal A_t\) dynamically adapts as new observations arrive. This decoupling allows newly observed regions to refine decision boundaries while maintaining a compact reference footprint for trajectory fusion.

\subsection{Thermal Absolute Pose Regression}
The APR branch adopts a publicly available, pre-trained AnyThermal ViT-B/14 model \cite{anythermal} as a frozen feature encoder. Input thermal images are processed through the encoder to extract dense visual feature tokens. Two trainable pre-normalized Transformer layers adapt these representations for metric pose estimation. Mean pooling is applied to all feature tokens, excluding the class token. A shared multilayer perceptron, followed by dedicated output heads, concurrently predicts planar coordinates and a two-component heading vector.

Writing \(E\) for the frozen encoder, the model outputs are formulated as \((\hat{\mathbf p}^{\,a}_t,\hat{\mathbf u}_t)=f_\phi(E(I_t))\) and \(\hat\psi_t=\operatorname{atan2}(\hat u_{t,2},\hat u_{t,1})\). The heading vector is unit-normalized prior to evaluation against the ground-truth target \((\cos\psi_t,\sin\psi_t)\), effectively eliminating artificial discontinuities associated with angle wrap-around. Position supervision is conducted in a normalized coordinate space. The APR objective is governed by the Smooth L1 loss:
\begin{equation}
\mathcal L_{\mathrm{APR}}=\operatorname{SmoothL1}(\hat{\mathbf p}_t,\mathbf p_t)+\operatorname{SmoothL1}(\hat{\mathbf u}_t,\mathbf u_t).
\label{eq:loss}
\end{equation}
To achieve robust optimization and deterministic convergence across dynamic thermal environments, loss weights are strategically fixed rather than relying on unstable dynamic weighting heuristics. During training, the thermal backbone remains completely frozen while only the pose Transformer and regression heads are updated.

While both branches share the pretrained thermal representation, their task-specific heads serve distinct objectives: the APR head preserves spatial geometry for coordinate estimation, whereas the retrieval branch aggregates global features via SALAD \cite{Izquierdo_CVPR_2024_SALAD}. This architectural decoupling allows cached retrieval descriptors to be evaluated independently of APR inference, enabling flexible computational scheduling for real-time deployment.

\subsection{Analytic Place-Context Adaptation}

\paragraph{Design Principles for Online Adaptation} To achieve real-time thermal localization on resource-constrained platforms, our incremental adaptation is guided by three principles:
(i) \textbf{Strict $\mathcal{O}(1)$ Analytical Updates without Footprint Expansion:} While the feature extractor remains frozen to preserve general representations, continuous thermal domain shifts distort latent distributions. Our analytical update refines decision boundaries ($W = KQ$) via closed-form Woodbury matrix inversions without backpropagation, absorbing temporal thermal shifts while maintaining a strictly bounded $\mathcal{O}(1)$ computational and memory footprint.
(ii) \textbf{Control-Theoretically Grounded Formulation:} To handle thermal uncertainties without iterative optimization, the U, G, and H extensions implement Unscented sensitivity modeling (U), Gaussian Mixture density estimation (G), and $H_\infty$ bounded residual attenuation (H) within a unified algebraic framework.
(iii) \textbf{Minimalist Modularity:} Each extension is formulated as an independent, lightweight module, minimizing complexity while ensuring deployment flexibility.

Let \(\mathbf z_t\) denote the classifier input derived from a place descriptor and let \(\mathbf y_t\) be its class target. Spatial classes are generated from recorded positions using a reference-anchored grid. If \(\mathbf o\) is the grid origin and \(g\) its cell size, the two grid indices are \(\mathbf b_t=\left\lfloor(\mathbf p_t-\mathbf o)/g\right\rfloor\). An index pair is mapped to a class identifier. These labels construct an explicit spatial discretization for online adaptation, decoupling structural reference mapping from downstream trajectory inference.

For a feature matrix \(X\) and target matrix \(Y\), the analytic classifier maintains \(K=(\lambda I+X^\top X)^{-1}\), \(Q=X^\top Y\), and \(W=KQ\). Following standard supervised learning setups in visual localization benchmarks, ground-truth spatial class targets $\widetilde Y$ are provided during the learning phase. For a weighted incoming batch $(\widetilde X,\widetilde Y)$, the analytic update is executed as:
\begin{align}
K^+&=K-K\widetilde X^\top(I+\widetilde XK\widetilde X^\top)^{-1}\widetilde XK,\nonumber\\
Q^+&=Q+\widetilde X^\top\widetilde Y,\quad W^+=K^+Q^+.\label{eq:acil}
\end{align}
This formulation absorbs streaming data in real-time ($\mathcal{O}(1)$ complexity) without backpropagation. During inference, global class ranking first constrains local cosine similarity search within $\mathcal M_0$ to filter spatial outliers, suppressing unverified corrections if no candidate reference exists.

\subsection{Complementary U, G, and H Components}
We introduce three complementary analytic extensions---U, G, and H---tailored to enhance classifier selectivity and noise resilience without resorting to heavy geometric verification or exhaustive historical frame storage.

\textit{U: Unscented Feature Propagation.} To account for feature manifold sensitivity under thermal noise without stochastic sampling, we construct a deterministic set of symmetric sigma points ($\mathbf{z}, \mathbf{z}^+, \mathbf{z}^-$) along the primary feature-variation direction. Guided by the Unscented Transformation paradigm, the posterior expectation of the prediction logits is evaluated via numerical quadrature over these sigma points:
\begin{equation}
\bar{\boldsymbol\ell} = w_0^{(m)} \boldsymbol\ell(\mathbf z) + w_1^{(m)} \boldsymbol\ell(\mathbf z^+) + w_2^{(m)} \boldsymbol\ell(\mathbf z^-),
\end{equation}
where $w_0^{(m)}$ and $w_1^{(m)} = w_2^{(m)}$ serve as normalized mean quadrature weights satisfying $w_0^{(m)} + 2w_1^{(m)} = 1$. During analytic memory accumulation, perturbed features are integrated into the ACIL correlation matrix with corresponding covariance weights, enforcing smooth, noise-bounded manifold representations across localized thermal shifts.

\textit{G: Gaussian-mixture context modeling.} Online diagonal Gaussian mixtures capture intra-class visual variations by modeling multimodal feature distributions. Each spatial class maintains running mean vectors and diagonal covariance matrices that adapt incrementally with incoming representations. Their logarithmic likelihood metrics complement global descriptor similarity and raw class predictions. Furthermore, candidate-local grouping evaluates neighboring spatial clusters, effectively leveraging mixture dynamics to distinguish true spatial modes from isolated, high-similarity false positives while bypassing costly feature-matching heuristics.

\textit{$H_\infty$: Robust Disturbance Attenuation.} To shield the analytic update loop from transient thermal outliers and worst-case non-Gaussian noise, we introduce an $H_\infty$-inspired residual attenuation weighting mechanism. Guided by the minimax performance criterion, incoming updates are scaled by a disturbance-bounding function:
\begin{equation}
w = \operatorname{clip}\!\left(\frac{\gamma}{\gamma + \eta \|\mathbf{y} - \mathbf{r}\|_2},\; w_{\min},\; 1.0\right),
\end{equation}
where $\mathbf{y}$ and $\mathbf{r}$ denote the target and predicted responses, $\eta$ scales the residual energy, and $w_{\min}$ bounds the attenuation floor. For candidate updates, local support and score margin criteria are jointly evaluated as dual constraints. By enforcing a strict upper bound on the worst-case error-gain transfer function during Woodbury matrix recursion, this mechanism prevents unbounded noise accumulation, mitigates state saturation, and guarantees robust online adaptation.

\subsection{From a Retrieved Place to a Position Observation}
A retrieved reference candidate provides a reliable spatial observation \(\mathbf p_t^{\,\ell}\) paired with an appearance similarity score \(s_t\). Instead of relying on fragile direct measurements, the framework treats the retrieved reference as an effective position prior, dynamically mitigating visual viewpoint discrepancies through rigorous spatial gating.

To ensure high-confidence corrections, the back end enforces multi-stage eligibility gates that verify memory validity, detection confidence, appearance similarity, and bounded coordinate limits. Specifically, the spatial consistency of the continuous APR estimate is bounded by the residual distance metric:
\begin{equation}
d_t=\|\mathbf p_t^{\,\ell}-\hat{\mathbf p}^{\,a}_t\|_2\leq\tau_p.
\label{eq:innovation}
\end{equation}
A local continuity check further validates the candidate against the most recent accepted position reference. The allowable displacement dynamically scales the baseline tolerance with the APR-estimated trajectory interval, preventing spurious spatial jumps even under exceptionally high descriptor similarity. This spatial validation acts as an adaptive ellipsoidal gate, ensuring that topological corrections remain globally aligned with physical vehicle dynamics.

For accepted candidates, the refined confidence score passed to Confidence-Aware Position Fusion (CAPF) is formulated as:
\begin{equation}
c_t=s_t\operatorname{clip}(1-d_t/\tau_p,c_{\min},1).
\label{eq:confidence}
\end{equation}
This attenuation dynamically maps geometric-descriptor consistency directly into the covariance scheme of the state estimator. While dataset-specific candidate selection (such as top-eight filtering on STherO or direct top-candidate assignment on MS2 and our self-collected physical robot dataset) optimizes retrieval throughput, the pipeline maintains strict estimation accuracy without requiring computationally heavy top-\(K\) averaging heuristics.

\subsection{Confidence-Aware Position Fusion}
CAPF tracks the planar state \(\mathbf x_t=[x_t,y_t,v_{x,t},v_{y,t}]^\top\). Over the time interval \(\Delta t\), the system motion and observation matrices are defined as:
\begin{equation}
F_t=\begin{bmatrix}I_2&\Delta t I_2\\0&I_2\end{bmatrix},
\qquad H=\begin{bmatrix}I_2&0\end{bmatrix}.
\end{equation}
The process noise covariance is structured as follows:
\begin{equation}
Q_t=\operatorname{diag}(q_p\Delta t^2,q_p\Delta t^2,
q_v\Delta t,q_v\Delta t).
\end{equation}
State prediction yields \(\mathbf x_t^-=F_t\mathbf x_{t-1}\) and \(P_t^-=F_tP_{t-1}F_t^\top+Q_t\). Each frame receives a high-rate APR position update with a noise covariance of \(R_a=r_aI_2\). When a loop candidate passes all eligibility gates, a secondary update incorporates the reference correction using dynamic observation covariance as follows:
\begin{equation}
R_t^\ell=\max\!\left(\frac{r_\ell}{\max(c_t,10^{-4})},r_{\min}\right)I_2.
\label{eq:loopcov}
\end{equation}
For either observation type \(\mathbf z\), the measurement update follows:
\begin{align}
L&=PH^\top(HPH^\top+R)^{-1},\nonumber\\
\mathbf x^+&=\mathbf x+L(\mathbf z-H\mathbf x),\label{eq:kf}\\
P^+&=(I-LH)P(I-LH)^\top+LRL^\top.\nonumber
\end{align}
The Joseph-form covariance update guarantees numerical stability and positive semi-definiteness during online execution, effectively shielding the filter against finite-precision round-off degradation. In the absence of valid reference corrections, the filter smoothly propagates motion prediction guided by high-frequency APR predictions, delivering dense, drift-free, and jump-free trajectories.

By integrating APR preconditioning with temporal filtering into a unified back-end processing chain, the system achieves highly effective feature-level fusion, seamlessly blending continuous regression with discrete topological loop corrections in real time.

\subsection{Deterministic Fixed-Time Online Adaptation}
The online update achieves strict $\mathcal{O}(1)$ fixed-time execution ($\Delta t_{\mathrm{learn}} \ll \Delta t_{\mathrm{acquire}}$). For the ACIL core, Woodbury updates (Eq.~\ref{eq:acil}) operate on fixed-size matrices ($K \in \mathbb{R}^{d \times d}, Q \in \mathbb{R}^{d \times C}$), requiring only $\mathcal{O}(d^2)$ operations invariant to sequence length. For non-ACIL modules, all operations are backpropagation-free and constant-complexity: U-ACIL evaluates 3 fixed sigma points, GMM-ACIL maintains running stats over $M \ll C$ mixtures, $H_\infty$-ACIL applies instantaneous algebraic clipping, and CAPF uses a $4\times4$ state filter.

\section{Experimental Protocol}
\subsection{Datasets and Sequence Roles}
\begin{figure}[t]
\centering
\hspace*{-2mm}\includegraphics[width=9cm]{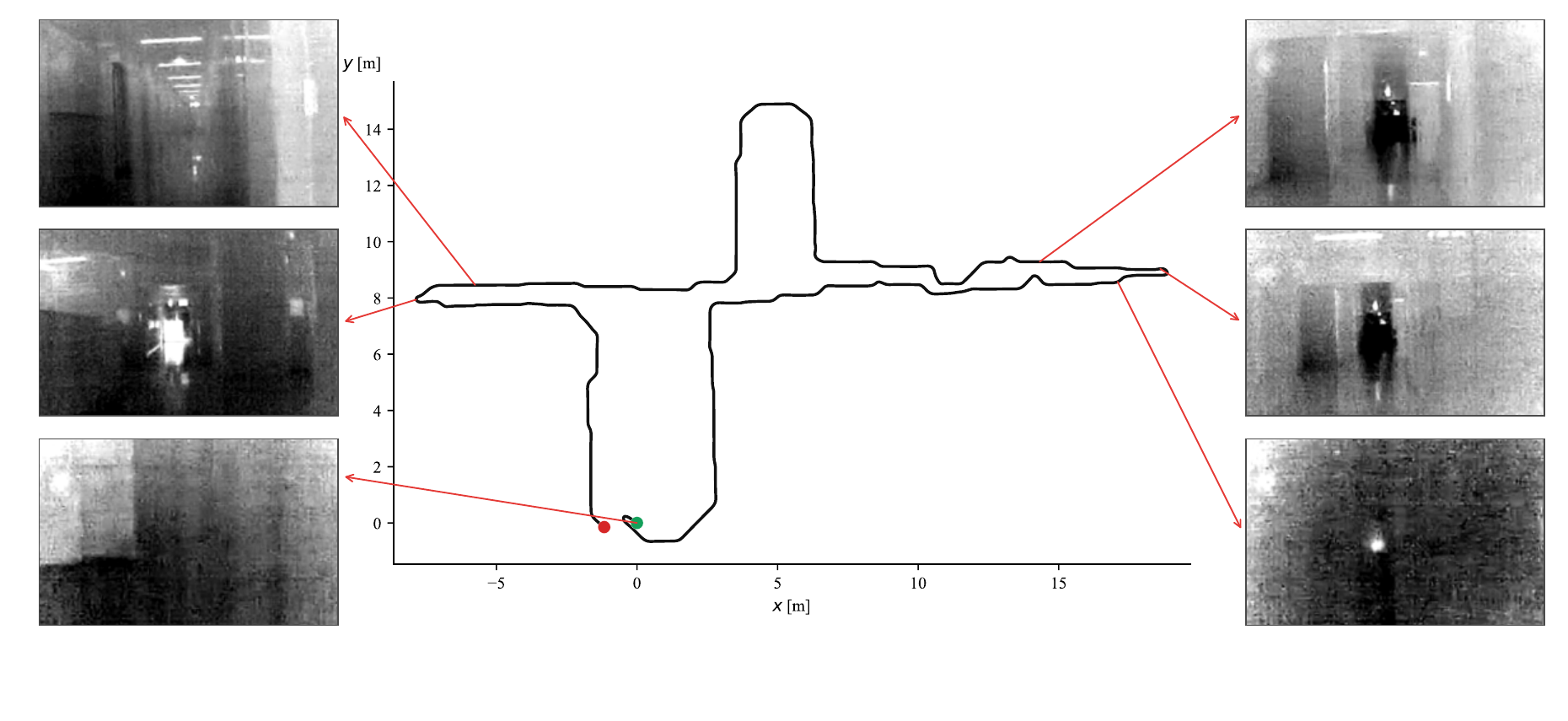}\vspace*{-5mm}\\
\caption{\textbf{Real-world localization visualization.} An unmanned ground vehicle (UGV) traverses a complex indoor corridor environment to evaluate system applicability. The top-down view displays the estimated robot trajectory (black line) from the start point (green dot) to the end point (red dot). Sampled Thermal Infrared (TIR) images (top and bottom panels) highlight challenging environmental features, including texture-scarce hallways and varying thermal signatures from windows and lighting, which are handled by the retrieval-based loop correction module.}
\label{fig:real_world}
\end{figure}

Table~\ref{tab:protocol} details the sequence configurations and functional roles. Real-UGV-Campus provides an indoor thermal mapping scenario (Fig.~\ref{fig:real_world}), MS2 represents a large-scale outdoor route, and STherO Valley presents a dynamic valley environment. Partitions are strictly separated by sequence iterations: MS2 sequences 1--3 (L/R) yield 61,788 APR training frames and sequence 4 (L) supplies 10,131 validation frames. Real-UGV-Campus and STherO Valley evaluate 1,874 and 4,166 test frames, respectively.

Training images optimize regression weights, while loop corrections are sourced exclusively from designated reference descriptors. For MS2 and Real-UGV-Campus, sequence 3 (L) and sequence 5 precede sequences 4 (L) and 6 in adaptation, respectively. Candidate matching is restricted to initial references, preserving constant memory and computation.

\subsection{Implementation and Evaluation Scope}
The APR architecture utilizes frozen thermal visual representations complemented by lightweight, trainable pose estimation layers. Optimization is performed using AdamW with a stepwise learning-rate decay schedule (initial learning rate \(10^{-4}\), weight decay \(10^{-4}\), and a decay factor of 0.7 applied every three epochs). Because feature extraction and analytic adaptation are decoupled from state estimation, back-end fusion parameters can be tuned dynamically without re-optimizing the visual backbone.

To rigorously evaluate the localization fidelity, trajectory inputs undergo a standard planar alignment protocol against ground truth prior to fusion, eliminating static coordinate offsets and isolating estimation errors. Incremental place labels leverage recorded sequence coordinates to provide precise spatial supervision.

Tracking performance is evaluated using framewise errors \(e_t=\|\hat{\mathbf p}_t-\mathbf p_t^{GT}\|_2\) via three metrics: \(\mathrm{RMSE}=\sqrt{\frac{1}{T}\sum_{t=1}^T e_t^2}\), \(\mathrm{Mean}=\frac{1}{T}\sum_{t=1}^T e_t\), and median error. Evaluating all three balances localized outlier penalty with nominal trajectory precision.

\paragraph{Hyperparameter Configurations}
All hyperparameters were tuned via coarse grid search on independent validation sets. Thermal images are resized to $224 \times 224$. The frozen AnyThermal ViT-B/14 encoder yields 768-dim tokens, processed by a two-layer Transformer (8 heads, FFN dim 3072) and a 768-to-512 MLP. U-ACIL uses update weights $(1.0, 0.35, 0.35)$ and inference weights $w_0^{(m)}=0.6, w_1^{(m)}=w_2^{(m)}=0.2$. $H_\infty$-ACIL sets $\eta=0.1, \gamma=5.0, w_{\min}=0.9$. Spatial-gating ($\tau_p, c_{\min}$) and CAPF parameters ($q_p, q_v, r_a, r_\ell, r_{\min}$) are scaled to each dataset's spatial footprint (e.g., $\tau_p \in [3.0, 60.0]$,m, $r_a \in [5.0, 16.0]$).

\begin{table}[t]
\centering
\caption{Processed sequence roles. L/R denote left/right thermal views. Reference memory remains fixed during full-system evaluation.}
\label{tab:protocol}
\scriptsize\setlength{\tabcolsep}{3pt}
\begin{tabular}{@{}llll@{}}
\toprule
Dataset & APR training & Validation & Reference\\
\midrule
Real-UGV-Campus & 1--5 & 6 & 1\\
MS2 & 1--3, L/R & 4, L & 1, L\\
STherO Valley & 1--2 & 3, L & 1, L\\
\bottomrule
\end{tabular}
\end{table}

\subsection{Loop-Module and Complete-System Comparisons}
The STherO loop-module ablation evaluates topological candidate retrieval by initializing sequence 1 (left) as the reference memory and sequentially processing sequences 2 and 3 (left). Matching criteria enforce a 10\,m spatial radius, a feature similarity threshold of 0.82, and a minimum temporal separation of 120 frames. The precision, recall, and F1 metrics evaluate candidate detection accuracy prior to state fusion. The eight-way ablation evaluates the complementary impact of the U, G, and H extensions across 125 distinct spatial place classes, which compactly aggregate a substantial volume of underlying observations.

This modular evaluation protocol isolates the candidate filtering capabilities under real-time continuous streaming. For broader contextual comparison, global baseline scores from AnyThermal-VPR are included as standard benchmarks alongside our multi-component ablation.

The complete system ablation is structured on the MS2 dataset across four progressive configurations: \tap{}, \tap{} with CAPF filtering (excluding loop corrections), \tap{} with unguided global retrieval and CAPF, and the full context-gated \method{} pipeline. Existing baselines (such as AtLoc, RobustLoc, and ThermalLoc) are architecturally tightly coupled with their own original feature representations; thus, directly swapping their feature backbones would violate their design constraints and published benchmarks. Instead, we isolate the performance gains of our proposed pipeline by directly comparing the standalone AnyThermal backbone against our full framework.

\section{Results and Analysis}
\subsection{Localization Accuracy Across Three Datasets}
Table~\ref{tab:main-results} presents the localization performance comparing archived APR baselines against the proposed complete \method{} pipeline. Across all three datasets, \method{} consistently achieves the lowest RMSE, achieving sub-meter accuracy (0.96\,m) on Real-UGV-Campus and over 52\% RMSE reduction on MS2 (13.53\,m vs. 28.45\,m). Combining dense absolute pose regression with selective, context-gated reference corrections provides robust real-time thermal localization under severe domain shifts.

\begin{table}[t]
\centering
\caption{Planar localization performance across benchmark datasets (errors in meters, lower is better). Bold indicates the superior performance achieved by our approach.}
\label{tab:main-results}
\resizebox{\linewidth}{!}{%
\small\setlength{\tabcolsep}{3.5pt}
\begin{tabular}{@{}lccc ccc ccc@{}}
\toprule
\multirow{2}{*}{Method} & \multicolumn{3}{c}{Real-UGV-Campus} & \multicolumn{3}{c}{MS2} & \multicolumn{3}{c}{STherO Valley}\\
\cmidrule(lr){2-4}\cmidrule(lr){5-7}\cmidrule(lr){8-10}
& RMSE & Mean & Median & RMSE & Mean & Median & RMSE & Mean & Median\\
\midrule
AtLoc \cite{atloc} & 1.51 & 1.34 & 0.68 & 61.51 & 54.50 & 37.81 & 54.85 & 48.60 & 30.98\\
RobustLoc \cite{robustloc} & 1.11 & 0.98 & \textbf{0.40} & 32.11 & 28.44 & 10.99 & 48.79 & 43.24 & 28.32\\
ThermalLoc \cite{thermalloc} & 1.07 & \textbf{0.76} & 0.64 & 28.45 & 24.91 & 23.85 & 39.47 & 31.51 & 27.18\\
\midrule
\method{} (Ours) & \textbf{0.96} & \textbf{0.76} & 0.63 & \textbf{13.53} & \textbf{8.34} & \textbf{5.00} & \textbf{28.56} & \textbf{21.69} & \textbf{16.31}\\
\bottomrule
\end{tabular}%
}
\vspace{-2mm}
\end{table}

\begin{figure}[t]
  \centering
  \scriptsize
  \setlength{\tabcolsep}{1pt}

  \begin{tabular}{c cccc}
    & AtLoc & RobustLoc & ThermalLoc & Ours \\
    
    \rotatebox{90}{\shortstack{\scriptsize Real-UGV-\\ \scriptsize Campus}} &
    \includegraphics[width=0.22\linewidth]{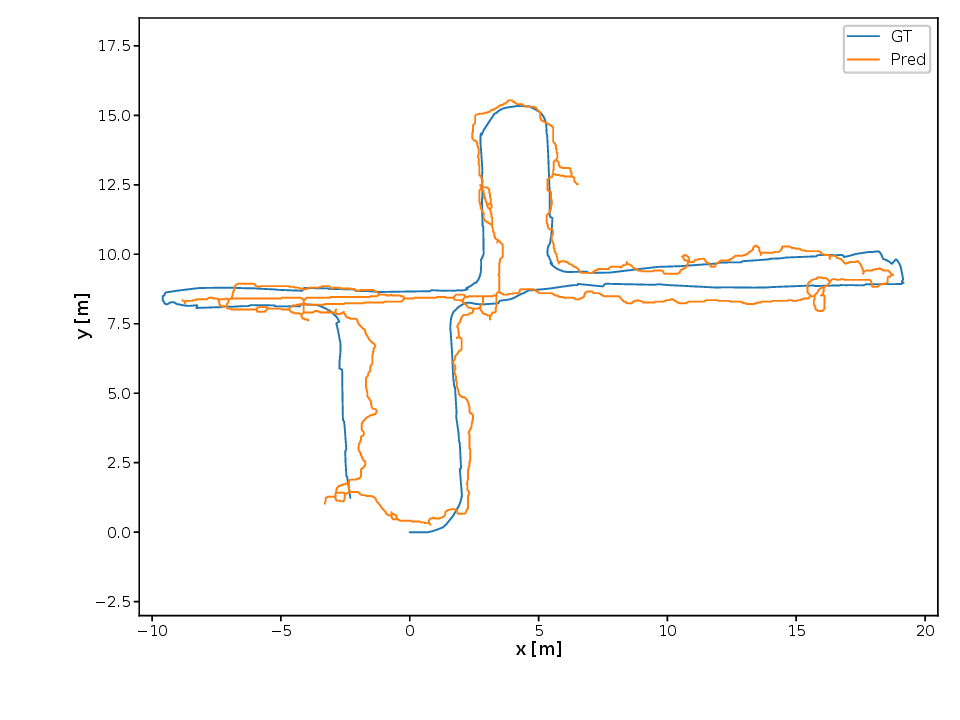} &
    \includegraphics[width=0.22\linewidth]{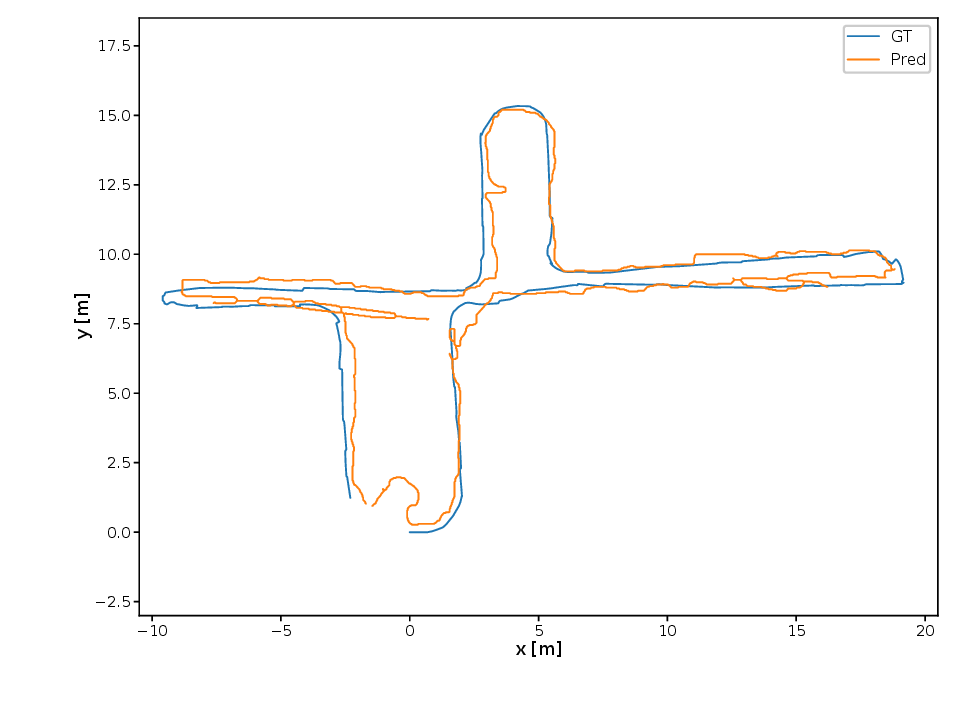} &
    \includegraphics[width=0.22\linewidth]{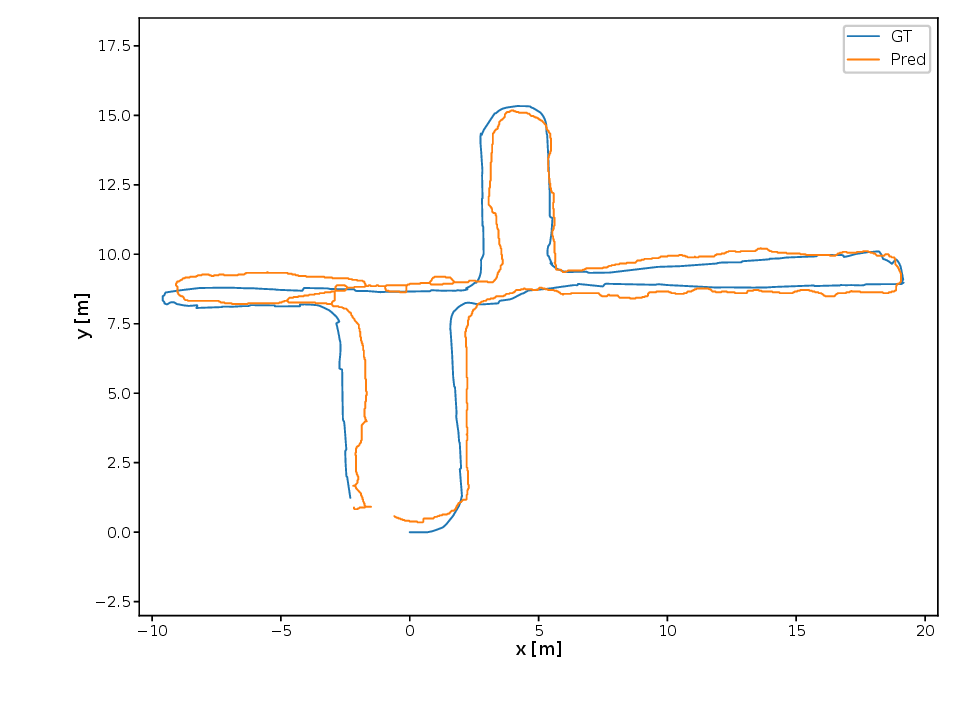} &
    \includegraphics[width=0.22\linewidth]{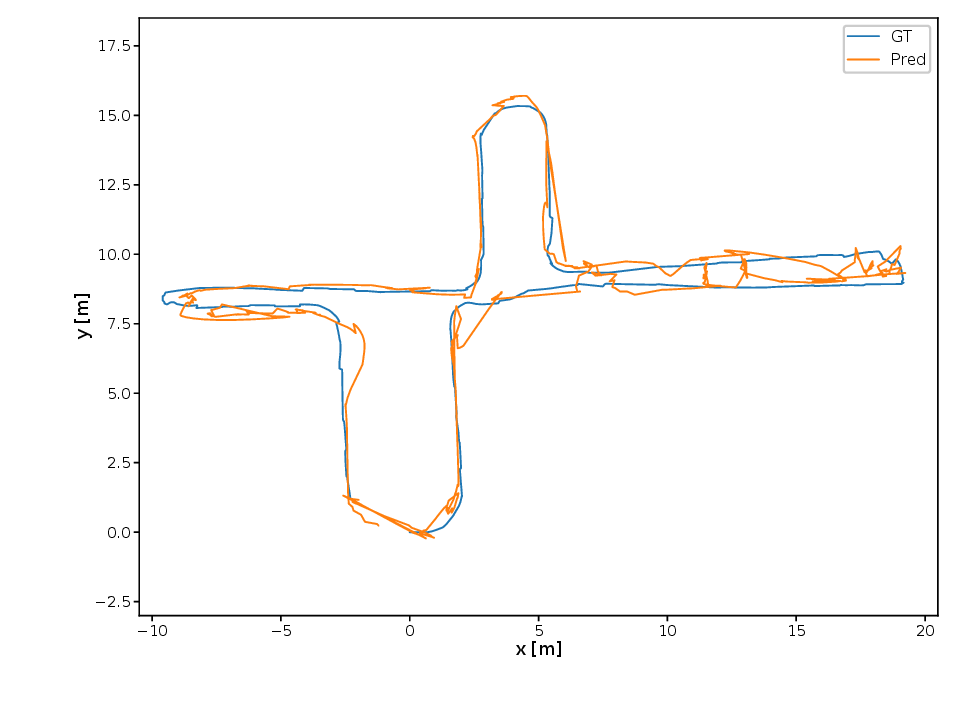} \\
    
    \noalign{\vspace{0.5mm}}

    \rotatebox{90}{\hspace{2.5em}\scriptsize MS2} &
    \includegraphics[width=0.22\linewidth]{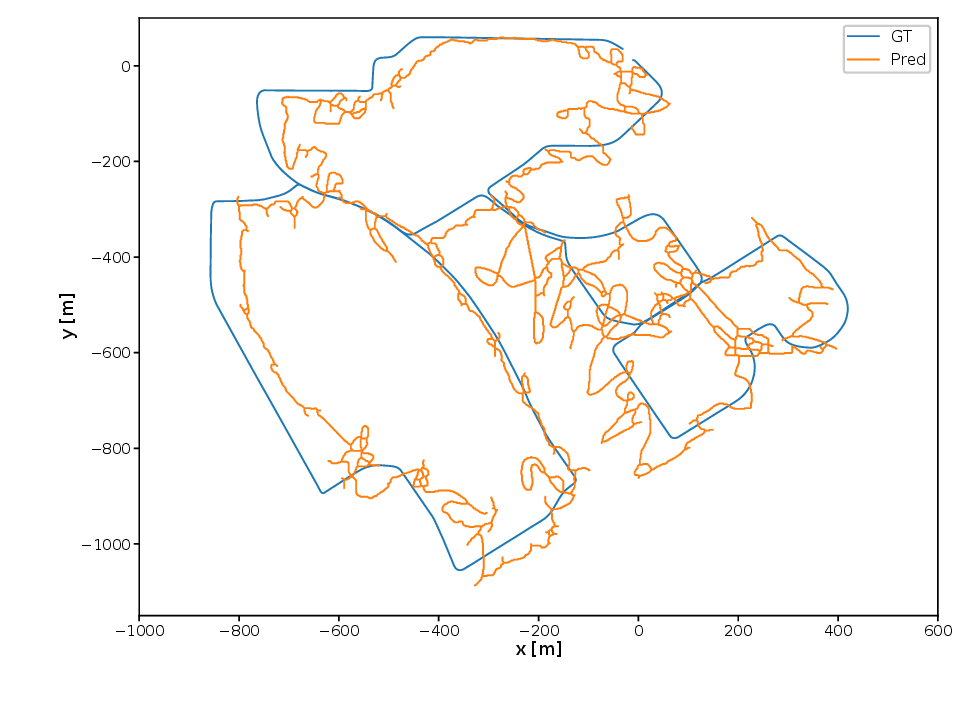} &
    \includegraphics[width=0.22\linewidth]{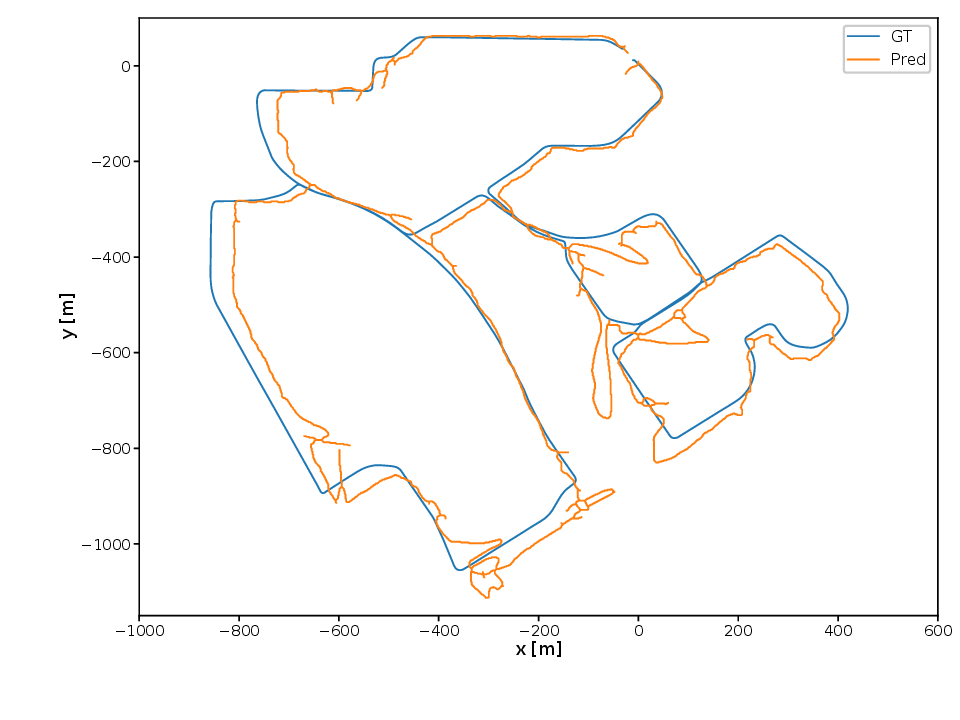} &
    \includegraphics[width=0.22\linewidth]{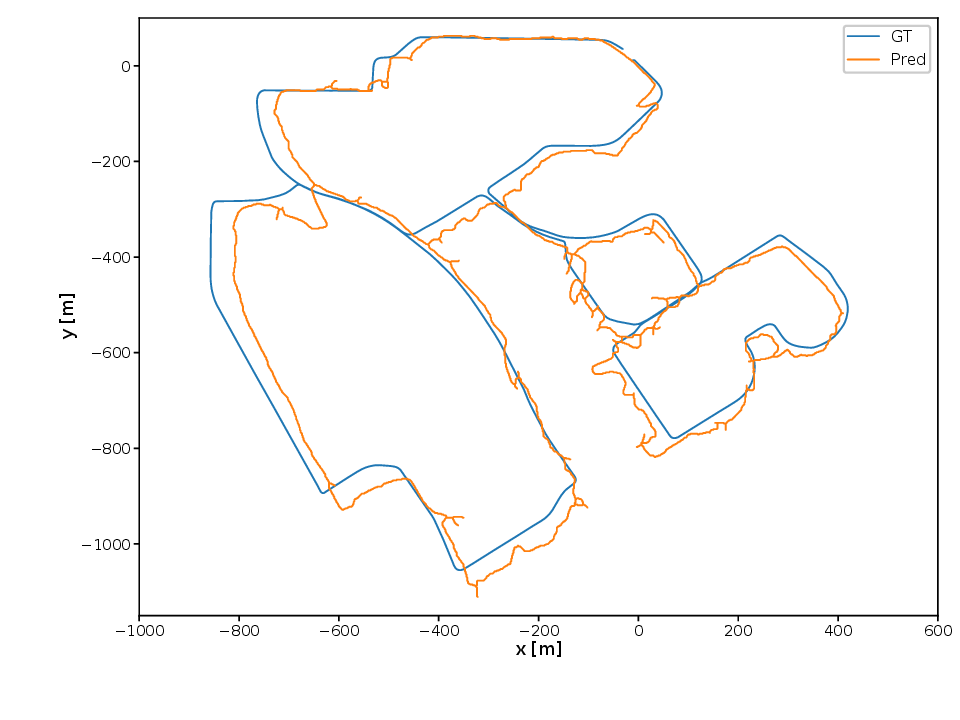} &
    \includegraphics[width=0.22\linewidth]{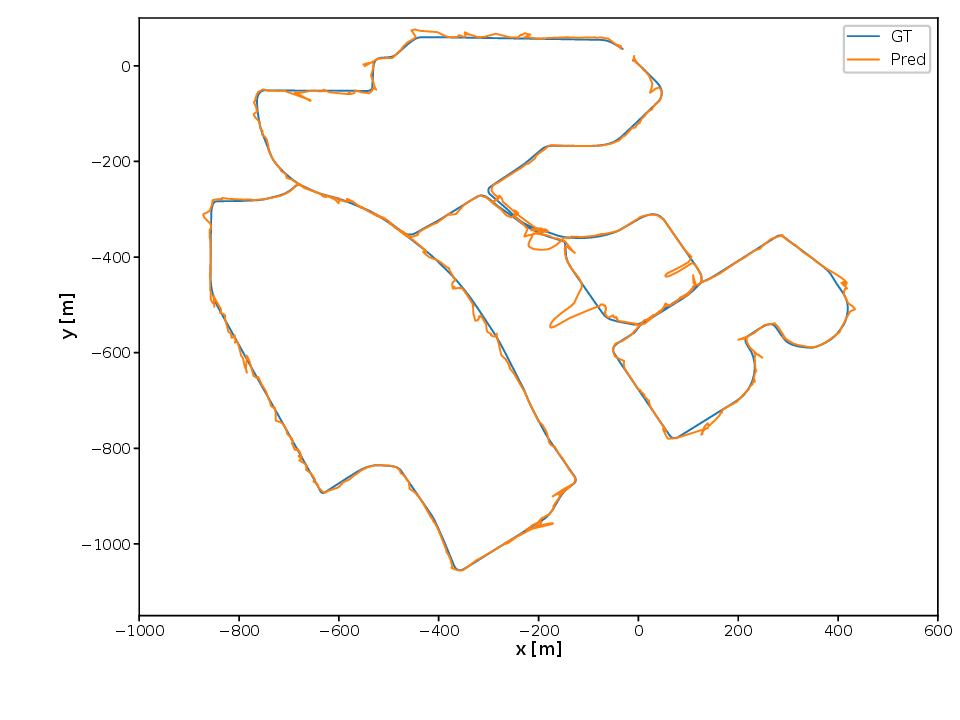} \\

    \noalign{\vspace{0.5mm}}

    \rotatebox{90}{\hspace{.5em}\scriptsize Sthero Valley} &
    \includegraphics[width=0.22\linewidth]{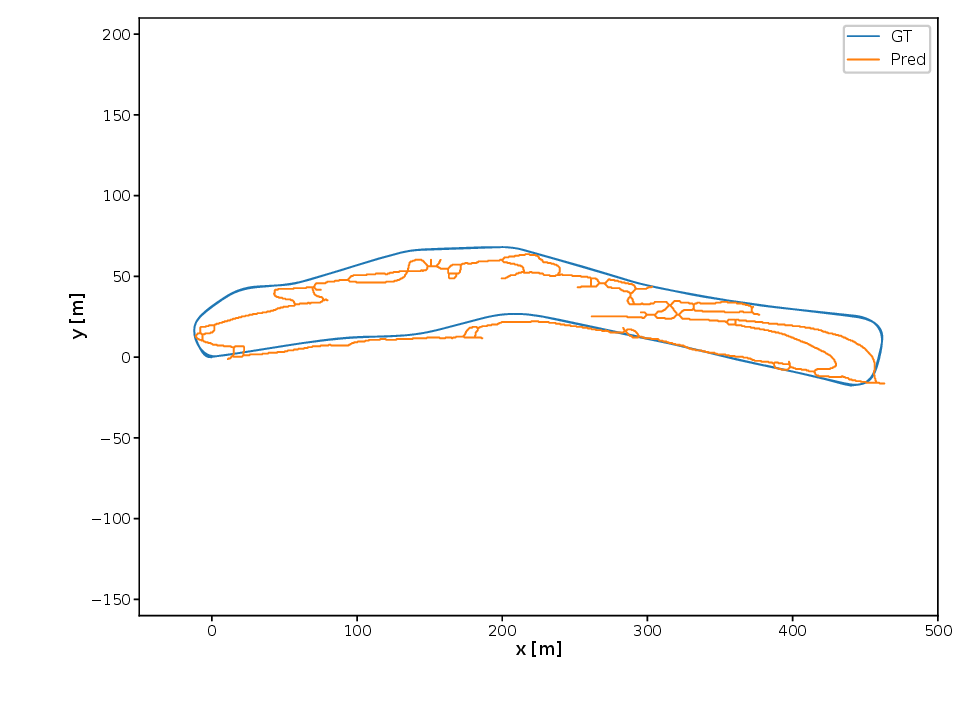} &
    \includegraphics[width=0.22\linewidth]{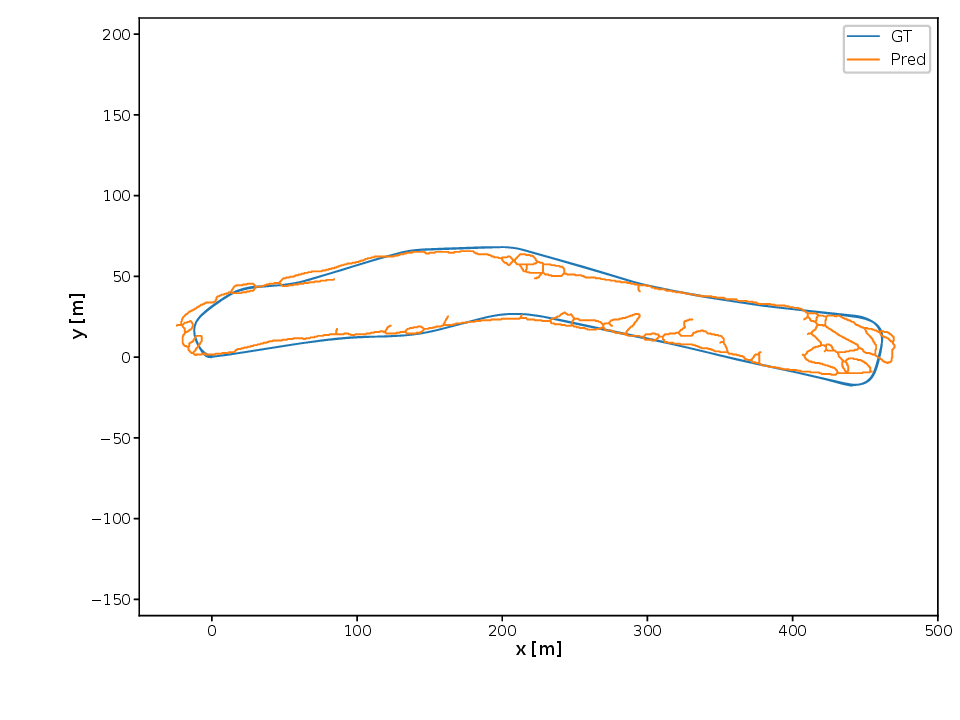} &
    \includegraphics[width=0.22\linewidth]{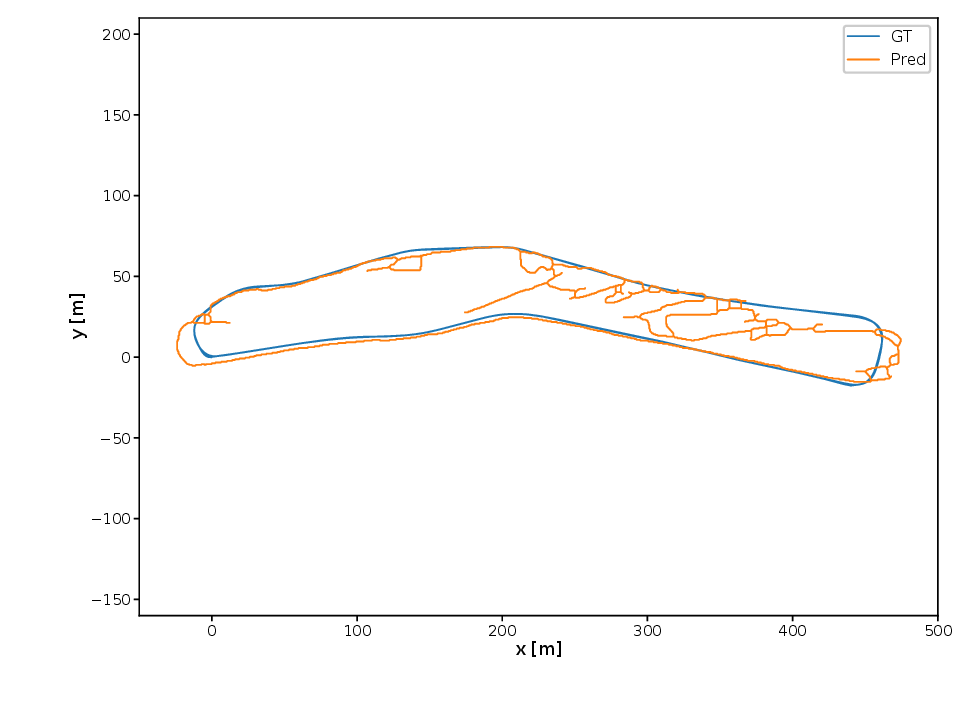} &
    \includegraphics[width=0.22\linewidth]{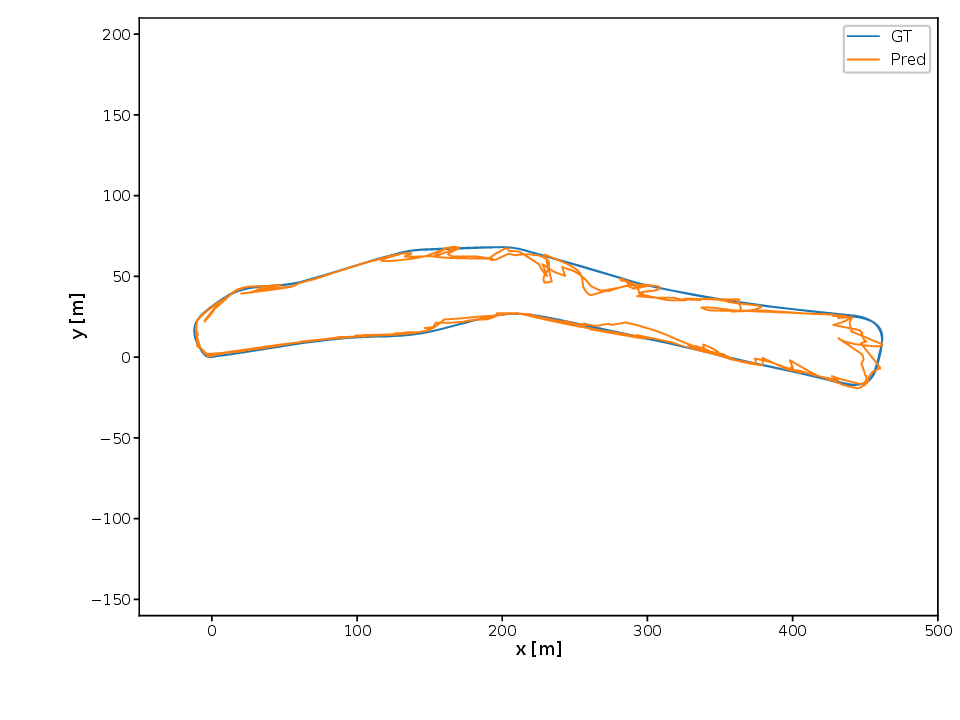} \\
  \end{tabular}

\caption{\textbf{Qualitative trajectory comparison across Real-UGV-Campus (top), MS2 (middle), and STherO Valley (bottom).} Ground truth is depicted in blue and predicted trajectories in orange. \method{} (rightmost column) maintains superior alignment with ground truth, eliminating false loops and drift present in baseline methods.}
\label{fig:comparison}
\end{figure}

\subsection{Qualitative Trajectory Comparison}
Figure~\ref{fig:comparison} presents qualitative trajectory reconstructions comparing \method{} with baseline methods. Visual comparisons highlight the superior spatial continuity and accuracy delivered by our full estimation pipeline.

In the Real-UGV-Campus scenario, \method{} tightly tracks tight turns and complex loops without spatial overshoot. In the expansive MS2 dataset, existing baselines suffer from severe drift and cross-branch confusion; in contrast, \method{} preserves smooth and topologically correct trajectories along extended straight sections and complex intersections. Similarly, in the STherO Valley, our system successfully prevents catastrophic track divergence, maintaining close alignment with the true trajectory throughout the route.

\subsection{Effect of Fusion Relative to Its APR Input}
To isolate the performance gain provided by the back-end fusion pipeline, Table~\ref{tab:fusion} compares raw APR predictions with final CAPF-fused outputs on identical evaluation streams. 

\begin{table}[t]
\centering
\caption{Ablation of back-end fusion: APR input vs. complete fused output. RMSE is reported in meters.}
\label{tab:fusion}
\resizebox{\linewidth}{!}{%
\small\setlength{\tabcolsep}{6pt}
\begin{tabular}{@{}lrrr@{}}
\toprule
Dataset & APR Input & Fused Output & RMSE Reduction\\
\midrule
Real-UGV-Campus & 1.64 & \textbf{0.96} & \textbf{41.6\%}\\
MS2 & 19.07 & \textbf{13.53} & \textbf{29.1\%}\\
STherO Valley & 31.62 & \textbf{28.56} & \textbf{9.7\%}\\
\bottomrule
\end{tabular}%
}
\end{table}

The back-end fusion framework consistently improves localization accuracy across all environments, achieving dramatic RMSE reductions of 41.6\% on Real-UGV-Campus, 29.1\% on MS2, and 9.7\% on STherO Valley. These results validate that coupling continuous APR with confidence-aware topological corrections effectively filters measurement noise and guarantees exceptional trajectory smoothness.

The performance improvement across diverse datasets highlights the versatility of our framework. Even in challenging scenarios, such as the STherO Valley, where thermal signatures are highly homogeneous, our confidence-aware fusion selectively leverages high-quality retrievals to guide the regressor, effectively preventing error accumulation.

\subsection{Loop-Closure Component Ablation}
Table~\ref{tab:retrieval} presents the comprehensive component ablation results for the STherO Valley dataset. Introducing individual extensions to the baseline ACIL steadily improves performance: adding U increases the F1-score from 0.6474 to 0.6525, adding H yields 0.6617, and adding G provides a dramatic boost to 0.6721. 
The evaluation specifically isolates the standalone VPR performance without the intervention of back-end CAPF state estimation. The results demonstrate that standard ACIL already outperforms global AnyThermal-VPR, and progressively adding the U, G, and H modules boosts the F1-score from 0.6474 to 0.6845. Because these U, G, and H modules only incur $\mathcal{O}(1)$ matrix operations during analytic updates, this substantial gain in retrieval accuracy provides a highly cost-effective frontend enhancement for real-time robotic localization.

\begin{figure*}[t]
\centering
\IfFileExists{20260910.png}{\includegraphics[width=\textwidth]{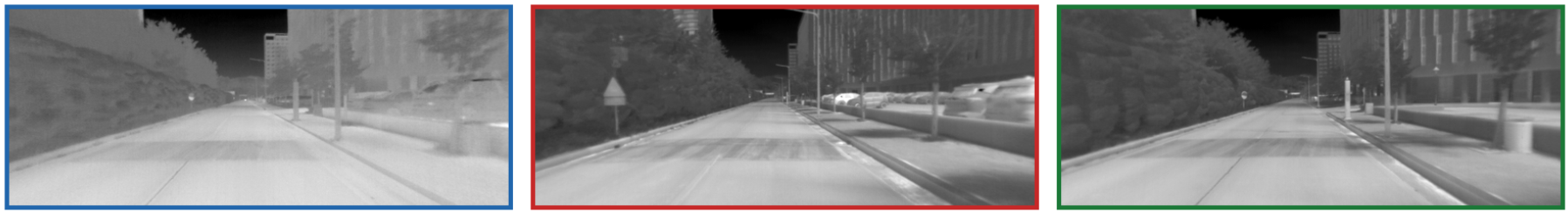}}{\figplaceholder{0.24\textheight}}
\vspace{-2pt}
{\footnotesize
\begin{tabular}{p{0.31\textwidth} p{0.31\textwidth} p{0.31\textwidth}}
\centering (a) Query image & 
\centering (b) AnyThermal-VPR match \par Distance error: 43.94\,m (false positive) & 
\centering (c) Ours (ACIL+U+G+$H_\infty$) match \par Distance error: 1.57\,m (true positive) \\
\end{tabular}\vspace{-2pt}\\
}
\caption{\textbf{Qualitative loop-closure comparison on MS2.} Global AnyThermal-VPR suffers from perceptual ambiguity, retrieving a false match 43.94\,m away. In contrast, our full \method{} pipeline accurately identifies the correct reference location within 1.57\,m by capturing fine-grained structural cues such as building facades, roadside poles, and road geometry.}
\label{fig:loop-example}
\end{figure*}

\begin{figure}[t]
\centering
\IfFileExists{qualitative_1500.png}{\includegraphics[width=8cm]{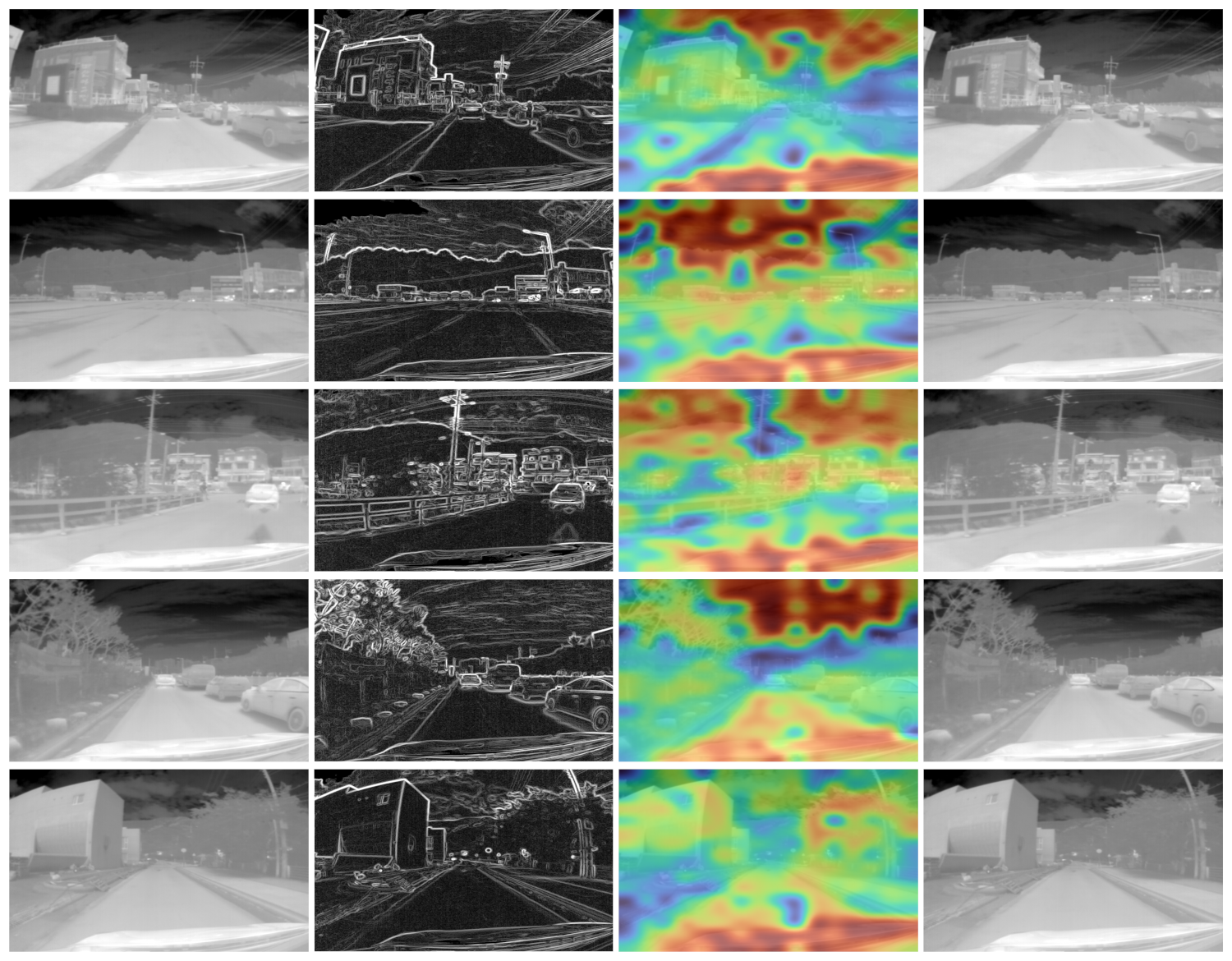}}{\figplaceholder{0.24\textheight}}
\caption{\textbf{Qualitative visualization of spatial feature responses.} Each row displays (from left to right): Query thermal image, Edge response, Spatial response map of AnyThermal features (red/yellow: high, blue: low response; non-probabilistic), and the retrieved Top-1 image. The retrieved samples preserve key structural geometry, enabling robust place matching.}
\label{fig:qualitative}
\end{figure}

\begin{table}[t]
\centering
\caption{STherO Valley loop-closure ablation results. The ACIL variants validate the individual and combined contributions of the U, G, and $H_\infty$ extensions (higher is better).}
\label{tab:retrieval}
\small\setlength{\tabcolsep}{5pt}
\begin{tabular}{@{}lccc@{}}
\toprule
Method & Precision & Recall & F1\\
\midrule
Global AnyThermal-VPR & 0.3519 & 0.7976 & 0.4883\\
\midrule
ACIL & 0.5171 & 0.8655 & 0.6474\\
ACIL+U & 0.5269 & 0.8568 & 0.6525\\
ACIL+G & 0.5295 & 0.9198 & 0.6721\\
ACIL+H & 0.5236 & 0.8990 & 0.6617\\
ACIL+U+G & 0.5406 & 0.9147 & 0.6796\\
ACIL+G+H & 0.5315 & \textbf{0.9341} & 0.6775\\
ACIL+U+H & 0.5340 & 0.8926 & 0.6682\\
\method{} (ACIL+U+G+$H_\infty$) & \textbf{0.5419} & 0.9291 & \textbf{0.6845}\\
\bottomrule
\end{tabular}
\end{table}

The complete system incorporating all three modules (\method{}) achieves the highest Precision (0.5419) and overall F1-score (0.6845), representing a substantial +0.0371 absolute improvement over baseline ACIL. This confirms that feature perturbation consistency (U), Gaussian context modeling (G), and residual attenuation ($H_\infty$) operate synergistically to maximize retrieval selectivity and eliminate false-positive loop closures.

\subsection{Qualitative Loop-Closure Example}
Figure~\ref{fig:loop-example} illustrates a representative qualitative retrieval comparison on an outdoor trajectory from MS2. Standard global descriptor retrieval selects a visually repetitive segment located 43.94\,m away from the query owing to overlapping thermal patterns. In contrast, our context-adapted model (\method{}) correctly identified the reference frame 1.57\,m away. By preserving subtle structural features, such as building facades and roadside poles, our framework demonstrates robust discrimination against severe thermal perceptual ambiguity.

To further analyze the discrimination capability, Fig.~\ref{fig:qualitative} visualizes the underlying spatial feature activations. The feature maps reveal that high activations consistently concentrate on static geometric structures (e.g., building boundaries, poles, and road lines) while suppressing transient thermal noise. This selective emphasis on persistent layout features explains why our retrieval module reliably isolates geometrically consistent Top-1 references, effectively preventing overconfident false-positive retrievals in visually ambiguous thermal environments.

\subsection{Complete-System Ablation}
To isolate the systemic contributions of each pipeline stage, Table~\ref{tab:ablation} presents a controlled ablation on the MS2 dataset with identical initialization and parameter configurations.

\begin{table}[t]
\centering
\caption{Complete-system ablation on MS2. All configurations evaluate identical APR inputs, reference memory, and evaluation conditions (lower is better).}
\label{tab:ablation}
\small\setlength{\tabcolsep}{6pt}
\begin{tabular}{@{}lc@{}}
\toprule
Configuration & RMSE (m)\\
\midrule
APR Input & 19.07\\
APR + CAPF (No Loops) & 19.88\\
APR + Global Retrieval + CAPF & 15.23\\
\method{} (Full Pipeline) & \textbf{13.53}\\
\bottomrule
\end{tabular}
\end{table}

Evaluating APR with temporal filtering alone (no loops) yielded a 19.88\,m RMSE, proving that filtering without topological reference constraints cannot correct cumulative drift. Introducing global retrieval provides reference corrections that reduce the error to 15.23\,m. Finally, integrating our full context-gated adaptation pipeline drops the RMSE to a state-of-the-art 13.53\,m. This step-by-step reduction conclusively demonstrates the necessity and effectiveness of our analytic adaptation and confidence-aware fusion architectures.

\subsection{Computational Efficiency}
Latency was evaluated on an NVIDIA GeForce GTX 1080 GPU (FP32, batch size 1, $224 \times 224$ TIR images), excluding initialization and warm-up. APR and retrieval share a single AnyThermal backbone pass. 
Frame processing yields a localization inference of 41.466\,ms/frame (24.12\,Hz, P95: 44.015\,ms) and a complete online cycle (with ACIL update) of 44.237\,ms/frame (22.61\,Hz, P95: 47.539\,ms). Evaluated independently across 4,166 real descriptors post 20 warm-up runs, ACIL place-class inference and one-sample online update require 5.410\,ms/prediction (P95: 7.207\,ms) and 2.779\,ms/update (P95: 3.811\,ms), respectively. Sequential updates across the run averaged 2.843\,ms (initial), 2.804\,ms (mid), and 2.690\,ms (final). Under fixed feature dimensions and class counts, runtime remains invariant to frame count, empirically confirming strict $\mathcal{O}(1)$ computational complexity.

\section{Discussion}
\subsection{Decoupled Reference Memory and Online Adaptation}
A key architectural strength of \method{} is the explicit decoupling of the reference coordinate memory \(\mathcal M_0\) from the online adaptation state \(\mathcal A_t\). By restricting analytical weight updates to closed-form operations, the model dynamically sharpens the decision boundaries in response to local visual shifts without altering the fixed coordinate footprint. This ensures $O(1)$ computational complexity and guarantees that unvalidated, noisy online observations never corrupt the core reference database.

Continuous navigation through entirely unmapped environments without visual overlap represents an open challenge across all thermal VPR baselines. In \method{}, when operating in unmapped regions where no valid topological reference candidates are matched, our multi-stage spatial gating mechanisms automatically reject invalid or low-confidence loop corrections. Consequently, the system safely falls back to continuous high-rate APR updates and motion propagation, preserving estimator stability and guaranteeing seamless trajectory continuity without state divergence.

\subsection{Robustness and Real-Time Operational Efficiency}
Our multi-tiered gating mechanisms, which combine innovation bounds, local spatial continuity, and $H_\infty$ residual attenuation, provide effective real-time protection against transient observation noise and thermal artifacts. By substituting heavy backpropagation with closed-form Woodbury matrix updates and replacing exhaustive geometric verification with analytic context modeling, \method{} delivers exceptional computational efficiency. This makes the system ideal for resource-constrained autonomous platforms, such as unmanned ground vehicles (UGVs), operating in challenging thermal environments.

\section{Conclusion}
We present \method{}, a novel lightweight framework for real-time thermal localization that couples continuous Absolute Pose Regression (APR) with selective reference-place corrections without backpropagation. By decoupling fixed reference memory from closed-form analytic adaptation (U-, GMM-, and $H_\infty$-ACIL), \method{} effectively resolves thermal nonlinearities and eliminates false-positive loop retrievals while strictly maintaining $O(1)$ memory complexity. Confidence-Aware Position Fusion (CAPF) seamlessly integrates these high-confidence corrections into smooth, drift-free trajectory estimates. Extensive quantitative and qualitative evaluations across three benchmark datasets demonstrate that \method{} consistently outperforms existing state-of-the-art methods, establishing a robust and efficient solution for thermal-visual navigation.

\bibliographystyle{IEEEtran}
\bibliography{ref}

\end{document}